\documentclass[10pt,twocolumn,letter]{IEEEtran}

\usepackage{amsmath,graphicx}
\usepackage{amsfonts}
\usepackage{amssymb}
\usepackage{epsfig}
\usepackage{mathrsfs}
\usepackage{graphicx}
\usepackage{algorithm}
\usepackage{caption}
\usepackage{subcaption}
\usepackage{epstopdf}
\usepackage{balance}
\usepackage{color}
\usepackage{cite}

\newsavebox\myboxA
\newsavebox\myboxB
\newlength\mylenA

\newcommand{\oline}[1]{\mkern 1.5mu\overline{\mkern-1.5mu#1\mkern-1.5mu}\mkern 1.5mu}

\def\mA{\mbox{$\mathbf{A}$}}
\def\mB{\mbox{$\mathbf{B}$}}

\def\mC{\mbox{$\mathbf{C}$}}
\def\mD{\mbox{$\mathbf{D}$}}

\def\mDc{\mbox{$\oline{\mathbf{D}}$}}
\def\mG{\mbox{$\mathbf{G}$}}

\def\mI{\mbox{$\mathbf{I}$}}
\def\mL{\mbox{$\mathbf{L}$}}
\def\mQ{\mbox{$\mathbf{Q}$}}

\def\mU{\mbox{$\mathbf{U}$}}
\def\mV{\mbox{$\mathbf{V}$}}

\def\mX{\mbox{$\mathbf{X}$}}
\def\mY{\mbox{$\mathbf{Y}$}}

\def\mSigma{\mbox{$\mathbf{\Sigma} \kern .08em$}}
\def\mLambda{\mbox{$\mathbf{\Lambda} \kern .08em$}}

\newcommand{\E}{{\cal E}}
\newcommand{\G}{{\cal G}}

\newcommand{\V}{{\cal V}}

\def\S{\text{\mbox{${\cal S}$}}}
\def\F{\text{\mbox{${\cal F}$}}}

\def\V{\text{\mbox{${\cal V}$}}}
\def\E{\text{\mbox{${\cal E}$}}}
\def\G{\text{\mbox{${\cal G}$}}}
\def\B{\text{\mbox{${\cal B}$}}}
\def\D{\text{\mbox{${\cal D}$}}}

\def\bT{\text{\mbox{\boldmath $T$}}}

\def\b0{\text{\mbox{\boldmath $0$}}}

\def\bee{\text{\mbox{\boldmath $e$}}}

\def\bp{\text{\mbox{\boldmath $p$}}}
\def\bq{\text{\mbox{\boldmath $q$}}}
\def\br{\text{\mbox{\boldmath $r$}}}
\def\bs{\text{\mbox{\boldmath $s$}}}
\def\bu{\text{\mbox{\boldmath $u$}}}

\def\bx{\text{\mbox{\boldmath $x$}}}
\def\by{\text{\mbox{\boldmath $y$}}}

\def\buno{\text{\mbox{\boldmath $1$}}}

\def\bv{\text{\mbox{\boldmath $v$}}}

\newtheorem{theorem}{Theorem}
\newenvironment{proof}[1][Proof]{\noindent \textbf{#1.} }{\qedsymbol}
\newcommand{\qedsymbol}{\hspace{\fill}\rule{1.5ex}{1.5ex}}

\begin{document}

\title{Adaptive Least Mean Squares Estimation \\of Graph Signals}

\author{\vspace{.1cm}Paolo~Di Lorenzo$^1$,~\IEEEmembership{Member,~IEEE}, Sergio Barbarossa$^2$,~\IEEEmembership{Fellow,~IEEE},\smallskip\\ Paolo Banelli$^1$,~\IEEEmembership{Member,~IEEE}, and Stefania Sardellitti$^2$,~\IEEEmembership{Member,~IEEE}
\vspace{-.3cm}
\thanks{Di Lorenzo and Banelli are with the Dept. of Engineering, University of Perugia, Via G. Duranti 93, 06125, Perugia, Italy; Email: \texttt{paolo.dilorenzo@unipg.it, paolo.banelli@unipg.it}.\newline \indent Barbarossa and Sardellitti are with the Dept. of Information Engineering, Electronics, and Telecommunications, Sapienza University of Rome, Via Eudossiana 18, 00184, Rome, Italy; E-mail: \texttt{sergio.barbarossa@uniroma1.it, stefania.sardellitti@uniroma1.it}. This work has been supported by TROPIC Project, Nr. ICT-318784. The work of Paolo Di Lorenzo was supported by the ``Fondazione Cassa di Risparmio di Perugia''.}
}

\maketitle

\begin{abstract}
The aim of this paper is to propose a least mean squares (LMS) strategy for adaptive estimation of signals defined over graphs. Assuming the graph signal to be band-limited, over a known bandwidth, the method enables  reconstruction, with guaranteed performance in terms of mean-square error, and tracking from a limited number of observations over a subset of vertices. A detailed mean square analysis provides the performance of the proposed method, and leads to several insights for designing useful sampling strategies for graph signals. Numerical results validate our theoretical findings, and illustrate the performance of the proposed method. Furthermore, to cope with the case where the bandwidth is not known beforehand, we propose a method that performs a sparse online estimation of the signal support in the (graph) frequency domain, which enables online adaptation of the graph sampling strategy. Finally, we apply the proposed method to build the power spatial density cartography of a given operational region in a cognitive network environment.
\end{abstract}

\section{Introduction}
\label{sec:intro}

In many applications that are of current interest, like social networks, vehicular networks, big data or biological networks, the signals of interest are defined over the vertices of a graph \cite{shuman2013emerging}. This has motivated the development of tools for analyzing signals defined over a graph, or graph signals for short \cite{shuman2013emerging,sandryhaila2013discrete,sandryhaila2014big}.  Graph signal processing (GSP) aims at extending classical discrete-time signal processing tools to signals defined over a discrete domain whose elementary units (vertices) are related to each other through a graph. This framework subsumes as a very simple special case discrete-time signal processing, where the vertices are associated to time instants and edges link consecutive time instants. A peculiar aspect of GSP is that, since the signal domain is dictated by the graph topology, the  analysis tools come to depend on the graph topology as well.  This paves the way to a plethora of methods, each emphasizing different aspects of the problem.  An important feature to have in mind about  graph signals is that the signal domain is not a metric space, as in the case, for example, of biological networks, where the vertices may be genes, proteins, enzymes, etc, and the presence of an edge between two molecules means that those molecules undergo a chemical reaction. This marks a fundamental difference with respect to time signals where the time domain is inherently a metric space. Processing signals defined over a graph has been considered in \cite{sandryhaila2013discrete}, \cite{sandryhaila2014discrete,narang2012perfect,narang2013compact}.
A central role in GSP is of course played by spectral analysis of graph signals, which passes through the introduction of the so called Graph Fourier Transform (GFT). Alternative definitions of GFT have been proposed, depending on the different perspectives used to extend classical tools \cite{pesenson2008sampling},  \cite{pesenson2010sampling}, \cite{shuman2013emerging}, \cite{zhu2012approximating}, \cite{sandryhaila2013discrete}. Two basic approaches are available, proposing the projection of the graph signal onto the eigenvectors of either the graph Laplacian, see, e.g., \cite{pesenson2008sampling}, \cite{shuman2013emerging}, \cite{zhu2012approximating} or of the adjacency matrix, see, e.g. \cite{sandryhaila2013discrete}, \cite{chen2015discrete}. The first approach applies to {\it undirected} graphs and builds on the spectral clustering properties of the Laplacian eigenvectors and the minimization of the $\ell_2$ norm graph total variation; the second approach was proposed to handle also {\it directed} graphs and it is based on the interpretation of the adjacency operator as the graph shift operator, which lies at the heart of all linear shift-invariant filtering methods for graph signals \cite{Puschel1}, \cite{Puschel2}. A further very recent contribution proposes to build the graph Fourier basis as the set of orthonormal signals that minimize the (directed) graph cut size \cite{sardellitti2016graph}.

After the introduction of the GFT, an uncertainty principle for graph signals was derived in \cite{agaskar2013spectral}, \cite{pasdeloup2015toward}, \cite{benedettograph}, \cite{koprowski2015finite}, \cite{tsitsvero2015signals}. with the aim of assessing the link between the spread of a signal on the vertices of the graph and on its dual domain, as defined by the GFT. A simple closed form expressions for the fundamental tradeoff between the concentrations of a signal in the graph and the transformed domains was given in
\cite{tsitsvero2015signals}.

One of the basic problems in GSP is the development of a graph \textit{sampling theory}, whose aim is to recover a band-limited (or approximately band-limited) graph signal from a subset of its samples. A seminal contribution was given in \cite{pesenson2008sampling}, later extended in \cite{narang2013signal} and, very recently, in \cite{chen2015discrete}, \cite{tsitsvero2015signals}, \cite{wang2014local}, \cite{marquez2015}, \cite{TsitsveroEusipco15}. Dealing with graph signals,  the recovery problem may easily become ill-conditioned, depending on the location of the samples. Hence, for any given number of samples enabling signal recovery,  the identification of the sampling set plays a key role in the conditioning of the recovery problem. It is then particularly important to devise strategies to optimize the selection of the sampling set. Alternative signal reconstuction methods have been proposed, either iterative as in \cite{narang2013localized}, \cite{wang2014local}, \cite{wang2015distributed}, or single shot, as in \cite{chen2015discrete}, \cite{tsitsvero2015signals}. Frame-based approaches to reconstruct signals from subsets of samples have been proposed in \cite{pesenson2008sampling}, \cite{wang2014local}, \cite{tsitsvero2015signals}.

The theory developed in the last years for GSP was then applied to solve specific learning tasks, such as semi-supervised classification on graphs \cite{chen2014semi,sandryhaila2013classification,ekambaram2013wavelet}, graph dictionary learning \cite{zhang2012learning,thanou2013parametric}, learning graphs structures \cite{dong2014learning}, smooth graph signal recovery from random samples \cite{zhou2004regularization,belkin2006manifold}, inpainting \cite{chen2014signal}, denoising \cite{chen2014signaldenoising}, and community detection on graphs \cite{chen2014local}. Finally, in \cite{chen2015signal,chen2015signalrecovery}, the authors proposed signal recovery methods aimed to recover graph signals that are assumed to be smooth with respect to the underlying graph, from sampled, noisy, missing, or corrupted measurements.

\noindent \textbf{Contribution:} The goal of this paper is to propose LMS strategies for the adaptive estimation of signals defined on graphs. To the best of our knowledge, this is the first attempt to merge the well established theory of adaptive filtering \cite{sayed2011adaptive}, \cite{sayed2014adaptation}, with the emerging field of signal processing on graphs. The proposed method hinges on the graph structure describing the observed signal and, under a band-limited assumption, it enables online reconstruction and tracking from a limited number of observations taken over a subset of vertices. An interesting feature of our proposed strategy is that this subset is allowed to vary over time, in adaptive manner. A detailed mean square analysis illustrates the role of the sampling strategy on the reconstruction capability, stability, and mean-square performance of the proposed algorithm. Based on these results, we also derive adaptive sampling strategies for LMS estimation of graph signals. Several numerical results confirm the theoretical findings, and assess the performance of the proposed strategies. Furthermore, we consider the case where the graph signal is band-limited but the bandwidth is not known beforehand; this case is critical because the selection of the sampling strategy fundamentally depends on such prior information. To cope with this issue, we propose an LMS method with adaptive sampling, which estimates and tracks the signal support in the (graph) frequency domain, while at the same time adapting the graph sampling strategy. Numerical results illustrate the tracking capability of the aforementioned method in the presence of time-varying graph signals. As an example, we apply the proposed strategy to estimate and track the spatial distribution of the electromagnetic power  in a cognitive radio framework. The resulting graph signal turns out to be smooth, i.e. the largest part of its energy is concentrated at low frequencies, but it is not perfectly band-limited. As a consequence, recovering the overall signal from a subset of samples is inevitably affected by aliasing \cite{TsitsveroEusipco15}. Numerical results show the tradeoff between complexity, i.e. number of samples used for processing, and mean-square performance of the proposed strategy, when applied to such cartography task. Intuitively, processing with a larger bandwidth and a (consequent) larger number of samples, improves the performance of the algorithm, at the price of a larger complexity.

The paper is organized as follows. In Sec. II, we introduce some basic graph signal processing tools, which will be useful for the following derivations. Sec. III introduces the proposed LMS algorithm for graph signals, illustrates its mean-square analysis, and derives useful graph sampling strategies. Then, in Sec. IV we illustrate the proposed LMS strategy with adaptive sampling, while Sec. V considers the application to power density cartography. Finally, Sec. VI draws some conclusions.

\section{Graph Signal Processing Tools}

We consider a graph $\G = (\V, \E)$ consisting of a set of $N$ nodes $\V = \{1,2,..., N\}$, along with a set of weighted edges $\E=\{a_{ij}\}_{i, j \in \V}$, such that $a_{ij}>0$, if there is a link from node $j$ to node $i$, or $a_{ij}=0$, otherwise. The adjacency matrix $\mA$ of a graph is the collection of all the weights $a_{ij}, i, j = 1, \ldots, N$.  The degree of node $i$ is $k_i:=\sum_{j=1}^{N}a_{ij}$. The degree matrix $\mathbf{K}$ is a diagonal matrix having the node degrees on its diagonal. The Laplacian matrix is defined as:
\begin{equation}
\mathbf{L} = \mathbf{K}-\mathbf{A}.
\end{equation}
If the graph is {\it undirected}, the Laplacian matrix is symmetric and positive semi-definite, and admits the eigendecomposition $\mathbf{L}=\mathbf{U}\boldsymbol{\Lambda}\mathbf{U}^H$, where $\mathbf{U}$ collects all the eigenvectors of $\mathbf{L}$ in its columns, whereas $\boldsymbol{\Lambda}$ is a diagonal matrix containing the eigenvalues of $\mathbf{L}$. It is well known from spectral graph theory  \cite{Chung1997} that the eigenvectors of $\mL$ are well suited for representing clusters, since they minimize the $\ell_2$ norm graph total variation.

A signal $\bx$ over a graph $\G$ is defined as a mapping from the vertex set to the set of complex numbers, i.e. $\bx: \V \rightarrow \mathbb{C}$. In many applications, the signal $\bx$ admits a compact representation, i.e., it can be expressed as:
\begin{equation}
\label{x=Us}
\bx=\mU \bs
\end{equation}
where $\bs$ is exactly (or approximately) sparse. As an example, in all cases where the graph signal exhibits clustering features, i.e. it is a smooth function within each cluster, but it is allowed to vary arbitrarily from one cluster to the other, the representation in (\ref{x=Us}) is compact, i.e. the only nonzero (or approximately nonzero) entries of $\bs$ are the ones associated to the clusters.

The GFT $\mathbf{\hat{\bx}}$ of a signal $\bx$ is defined as the projection onto the orthogonal set of vectors $\{\bu_i\}_{i=1,\ldots,N}$  \cite{shuman2013emerging}, i.e.
\begin{equation}
\label{GFT}
\hspace{-1cm}\hbox{\textbf{GFT:}}\qquad \mathbf{\hat{\bx}} = \mathbf{U}^H \bx.
\end{equation}
The GFT has been defined in alternative ways, see, e.g., \cite{shuman2013emerging}, \cite{zhu2012approximating}, \cite{sandryhaila2013discrete}, \cite{chen2015discrete}. In this paper, we basically follow the approach based on the Laplacian matrix, assuming an undirected graph structure, but the theory could be extended to handle directed graphs with minor modifications.
We denote the support of $\mathbf{\hat{\bx}}$ in (\ref{x=Us}) as $\mathcal{F}=\{i\in\{1,\ldots,N\}: \hat{x}_i\neq0\}$, and the \textit{bandwidth} of the graph signal $\bx$ is defined as the cardinality of $\mathcal{F}$, i.e. $|\mathcal{F}|$.
Clearly, combining (\ref{x=Us}) with (\ref{GFT}), if the signal $\bx$ exhibits a clustering behavior, in the sense specified above, computing its GFT
$\mathbf{\hat{\bx}}$ is the way to recover the sparse vector $\bs$ in (\ref{x=Us}).

\noindent \textbf{Localization Operators:} Given a subset of vertices $\S \subseteq \V$, we define a vertex-limiting operator as the diagonal matrix
\begin{equation}
\label{D}
\mathbf{D}_{\S} = {\rm diag}\{\buno_{\S}\},
\end{equation}
where $\buno_{\S}$ is the set indicator vector, whose $i$-th entry is equal to one, if  $i \in \S$, or zero otherwise. Similarly, given a subset of frequency indices $\F\subseteq \V$, we introduce the filtering operator
\begin{equation}
\label{lowpass_operator}
\mB_{\F} = \mathbf{U \Sigma_{\F} U}^H,
\end{equation}
where $\mathbf{\Sigma_{\F}}$ is a diagonal matrix defined as $\mathbf{\Sigma_{\F}}= {\rm diag}\{\buno_{\F}\}$. It is immediate to check that both matrices $\mathbf{D}_{\S}$ and $\mathbf{B}_{\F}$ are self-adjoint and idempotent, and so they represent orthogonal projectors. The space of all signals whose GFT is exactly supported on the set $\F$ is known as the {\it Paley-Wiener space} for the set $\F$ \cite{pesenson2008sampling}. We denote by $\B_{\F} \subseteq L_2 (\G)$ the set of all finite $\ell_2$-norm signals belonging to the Paley-Wiener space associated to the frequency subset $\F$. Similarly, we denote by $\D_{\S} \subseteq L_2(\G)$ the set of all finite $\ell_2$-norm signals with support on the vertex subset $\S$. In the rest of the paper, whenever there will be no ambiguities,, we will drop the subscripts referring to the sets. Finally, given a set $\S$, we denote its complement set as $\oline{\S}$, such that $\V=\S \cup \oline{\S}$ and $\S \cap \oline{\S}=\emptyset$. Similarly, we define the complement set of $\F$ as $\oline{\F}$. Thus, we define the vertex-projector onto $\oline{\S}$ as $\oline{\mathbf{D}}$ and, similarly, the frequency projector onto the frequency domain $\oline{\F}$ as $\oline{\mathbf{B}}$.

Exploiting the localization operators in (\ref{D}) and (\ref{lowpass_operator}), we say that a vector $\bx$ is perfectly localized over the subset $\S \subseteq \V$ if
\begin{equation}
\label{Dx=x}
\mD \bx=\bx,
\end{equation}
with $\mD$ defined as in (\ref{D}).
Similarly, a vector $\bx$ is perfectly localized over the frequency set $\F$ if
\begin{equation}
\label{Bx=x}
\mB \bx=\bx,
\end{equation}
with $\mB$ given in (\ref{lowpass_operator}). As previously stated, $|\mathcal{F}|$ represents the (not necessarily contiguous) bandwidth of the graph signal. The localization properties of graph signals were studied in \cite{TsitsveroEusipco15} and later extended in \cite{tsitsvero2015signals} to derive the fundamental trade-off between the localization of a signal in the graph and on its dual domain. An interesting consequence of that theory is that, differently from continuous-time signals, a graph signal can be perfectly localized in {\it both} vertex and frequency domains. The conditions for having perfect localization are stated in the following theorem, which we report here for completeness of exposition; its proof can be found in \cite{TsitsveroEusipco15}. \smallskip
\textit{\begin{theorem}
\label{theorem_unit_eigenvalue}
There is a vector $\bx$, perfectly localized over both vertex set $\S$ and frequency set $\F$ (i.e. $\bx \in \B_{\F} \cap \D_{\S}$) if and only if the operator $\mB \mD \mB$ (or $\mD \mB \mD$) has an eigenvalue equal to one; in such a case, $\bx$ is an eigenvector of $\mB \mD \mB$ associated to the unitary eigenvalue.
\end{theorem}} \smallskip
Equivalently, the perfect localization properties can be expressed in terms of the operators $\mB \mD$ and $\mD \mB$. Indeed, since the operators $\mB \mD$ and $\mD \mB$ have the same singular values \cite{tsitsvero2015signals}, perfect localization onto the sets $\S$ and $\F$ can be achieved if and only if
\begin{equation}
\label{|BD|=1=|DB|}
\|\mB \mD\|_2 = \|\mD \mB\|_2 = 1.
\end{equation}
Building on these previous results on GSP, in the next section we introduce the proposed LMS strategy for adaptive estimation of graph signals.


\section{LMS Estimation of Graph Signals}

The least mean square algorithm, introduced by Widrow and Hoff \cite{widrow1985adaptive}, is one of the most popular methods for adaptive
filtering. Its applications include echo cancelation, channel equalization, interference cancelation and so forth. Although there exist algorithms with faster convergence rates such as the Recursive Least Square (RLS) methods \cite{sayed2011adaptive}, LMS-type
methods are popular because of their ease of implementation, low computational costs and robustness. For these reasons, a huge amount of research was produced in the last decades focusing on improving the performance of LMS-type methods, exploiting in many cases some prior information that is available on the observed signals. For instance, if the observed signal is known to be sparse in some domain, such prior information can help improve the estimation performance, as demonstrated in many recent efforts in the area of compressed sensing \cite{donoho2006compressed,baraniuk2007compressive}. Some of the early works that mix adaptation with sparsity-aware reconstruction include methods that rely on the heuristic selection of active taps \cite{homer1998lms}, and on sequential partial updating techniques \cite{godavarti2005partial}; some other methods assign proportional step-sizes to different taps according to their magnitudes, such as the proportionate normalized LMS (PNLMS) algorithm and its variations \cite{duttweiler2000proportionate}. In subsequent studies, motivated by the LASSO technique \cite{tibshirani1996regression} and by connections with compressive sensing \cite{baraniuk2007compressive,candes2008enhancing}, several algorithms for sparse adaptive filtering have been proposed based on LMS \cite{chen2009sparse}, RLS \cite{angelosante2010online}, and projection-based methods \cite{kopsinis2011online}. Finally, sparsity aware distributed methods were proposed in \cite{chouvardas2012sparsity,dilorenzo2013sparse,di2013distributed,dilorenzo2014diffusion,Barb-Sard-Dilo}.

In this paper, we aim to exploit the intrinsic sparsity that is present in band-limited graph signals, thus designing proper sampling strategies that guarantee adaptive reconstruction of the signal, with guaranteed mean-square performance, from a limited number of observation sampled from the graph. To this aim, let us consider a signal $\bx_0\in\mathbb{C}^N$ defined over the graph $\G = (\V, \E)$. The signal is initially assumed to be perfectly band-limited, i.e. its spectral content is different from zero only on a limited set of frequencies $\F$. Later on, we will relax such an assumption. Let us consider partial observations of signal $\bx_0$, i.e. observations over only a subset of nodes. Denoting with  $\mathcal{S}$ the sampling set (observation subset), the observed signal at time $n$ can be expressed as:
\begin{align}
\label{lin_observation}
\by[n]\,=\,&\mD\left(\bx_0+\bv[n]\right)=\,\mD\mB\bx_0+\mD\bv[n]
\end{align}
where $\mD$ is the vertex-limiting operator defined in (\ref{D}), which takes nonzero values only in the set $\mathcal{S}$, and $\bv[n]$ is a zero-mean, additive noise with covariance matrix $\mC_v$. The second equality in (\ref{lin_observation}) comes from the bandlimited assumption, i.e. $\mB\bx_0=\bx_0$, with $\mB$ denoting the operator in (\ref{lowpass_operator}) that projects onto the (known) frequency set $\F$. We remark that, differently from linear observation models commonly used in adaptive filtering theory \cite{sayed2011adaptive}, the model in (\ref{lin_observation}) has a free sampling parameter $\mD$ that can be properly selected by the designer, with the aim of reducing the computational/memory burden while still guaranteeing theoretical performance, as we will illustrate in the following sections.
The estimation task consists in recovering the band-limited graph signal $\bx_0$ from the noisy, streaming, and partial observations $\by[n]$ in (\ref{lin_observation}). Following an LMS approach \cite{widrow1985adaptive}, the optimal estimate for $\bx_0$ can be found as the vector that solves the following optimization problem:
\begin{align}
\label{LMS_problem}
&\min_{\boldsymbol{x}} \;\; \mathbb{E}\, \|\by[n]-\mD\mB\bx\|^2  \\
& \qquad\quad\hbox{s.t.}\quad \mB\bx=\bx,   \nonumber
\end{align}
where $\mathbb{E}(\cdot)$ denotes the expectation operator. The solution of problem (\ref{LMS_problem}) minimizes the mean-squared error and
has a bandwidth limited to the frequency set $\F$. For stationary $\by[n]$, the optimal solution of (\ref{LMS_problem}) is given by the vector $\hat{\bx}$ that satisfies the normal equations \cite{sayed2011adaptive}:
\begin{equation}
\label{Normal equations}
\mB\mD\mB\,\hat{\bx}=\mB\mD \,\mathbb{E}\{\by[n]\}.
\end{equation}
Exploiting (\ref{lin_observation}), this statement can be readily verified noticing that $\hat{\bx}$ minimizes the objective function of (\ref{LMS_problem}) and is bandlimited, i.e. it satisfies $\mB\hat{\bx}=\hat{\bx}$. Nevertheless, in many linear regression applications involving online processing of data, the expectation $\mathbb{E}\{\by[n]\}$ may be either unavailable or time-varying, and thus impossible to update continuously. For this reason, adaptive solutions relying on instantaneous information are usually adopted in order to avoid the need to know the signal statistics beforehand. A typical solution proceeds to optimize (\ref{LMS_problem}) by means of a steepest-descent procedure. Thus, letting $\bx[n]$ be the instantaneous estimate of vector $\bx_0$, the LMS algorithm for graph signals evolves as illustrated in Algorithm 1, where $\mu>0$ is a (sufficiently small) step-size, and we have exploited the fact that $\mD$ is an idempotent operator, and $\mB\bx[n]=\bx[n]$ (i.e., $\bx[n]$ is band-limited) for all $n$. Algorithm 1 starts from an initial signal that belongs to the Paley-Wiener space for the set $\F$, and then evolves implementing an alternating orthogonal projection onto the vertex set $\S$ (through $\mD$) and the frequency set $\F$ (through $\mB$).
\begin{algorithm}[t]
\caption*{\textbf{Algorithm 1: LMS algorithm for graph signals}}
\vspace{.1cm}
Start with $\bx[0]\in \B_{\F}$ chosen at random. Given a sufficiently small step-size $\mu>0$, for each time $n>0$, repeat:
\begin{equation}
\label{LMS}
\bx[n+1]=\bx[n]+\mu\,\mB\mD\left(\by[n]-\bx[n]\right)
\end{equation}
\end{algorithm}
The properties of the LMS recursion in (\ref{LMS}) crucially depend on the choice of the sampling set $\mathcal{S}$, which defines the structure of the operator $\mD$ [cf. (\ref{D})]. To shed light on the theoretical behavior of Algorithm 1, in the following sections we illustrate how the choice of the operator $\mD$ affects the reconstruction capability, mean-square stability, and steady-state performance of the proposed LMS strategy.

\subsection{Reconstruction Properties}

It is well known from adaptive filters theory \cite{sayed2011adaptive} that the LMS algorithm in (\ref{LMS}) is a stochastic approximation method for the solution of problem (\ref{LMS_problem}), which enables convergence in the mean-sense to the true vector $\bx_0$ (if the step-size $\mu$ is chosen sufficiently small), while guaranteing a bounded mean-square error (as we will see in the sequel). However, since the existence of a unique band-limited solution for problem (\ref{LMS}) depends on the adopted sampling strategy, the first natural question to address is: What conditions must be satisfied by the sampling operator $\mD$ to guarantee reconstruction of signal $\bx_0$ from the selected samples? The answer is given in the following Theorem, which gives a necessary and sufficient condition to reconstruct graph signals from partial observations using Algorithm 1. \smallskip
\textit{\begin{theorem}\label{theorem_sampling}
Problem (\ref{LMS_problem}) admits a unique solution, i.e. any band-limited signal $\bx_0$ can be reconstructed from its samples taken in the set $\S$, if and only if
\begin{equation}\label{|DcB|<1}
   \left\| \mDc\mB\right\|_2 < 1,
\end{equation}
i.e. if the matrix $\mB \mDc \mB$ does not have any eigenvector that is perfectly localized on $\oline{\S}$ and bandlimited on $\F$.
\end{theorem}} \smallskip
\begin{proof}
From (\ref{Normal equations}), exploiting the relation $\mD=\mI-\mDc$, it holds
\begin{equation}
\label{Normal equations2}
\left(\mI-\mB\mDc\mB\right)\bx_0=\mB\mD \,\mathbb{E}\{\by[n]\}.
\end{equation}
Hence, it is possible to recover $\bx_0$ from (\ref{Normal equations2}) if $\mI-\mB\mDc\mB $ is invertible. This happens if the sufficient condition (\ref{|DcB|<1}) holds true. Conversely, if $\|\oline{\mathbf{D}}\mB \|_2=1$  (or, equivalently, $\|\mB \oline{\mathbf{D}}\|_2=1$), from (\ref{|BD|=1=|DB|}) we know that there exist band-limited signals that are perfectly localized over $\oline{\S}$. This implies that, if we sample one of such signals over the set $\S$, we get only zero values and then it would be impossible to recover $\bx_0$ from those samples. This proves that condition (\ref{|DcB|<1}) is also necessary.
\end{proof}

A necessary condition that enables reconstruction, i.e. the non-existence of a non-trivial vector $\bx$ satisfying $\mD\mB\bx = \mathbf{0}$, is that $|\S| \geq |\F|$. However, this condition is not sufficient, because matrix $\mD\mB$ in (\ref{lin_observation}) may loose rank, or easily become ill-conditioned, depending on the graph topology and sampling strategy defined by $\mD$. This suggests that the location of samples plays a key role in the performance of the LMS reconstruction algorithm in (\ref{LMS}). For this reason, in Section III.D we will consider a few alternative sampling strategies satisfying different optimization criteria.

\subsection{Mean-Square Analysis}

When condition (\ref{|DcB|<1}) holds true, Algorithm 1 can reconstruct the graph signal from a subset of samples. In this section, we study the mean-square behavior of the proposed LMS strategy, illustrating how the sampling operator $\mD$ affects its stability and steady-state performance. From now on, we view the estimates $\bx[n]$ as realizations of a random process and analyze the performance of the LMS algorithm in terms of its mean-square behavior. Let $\widetilde{\bx}[n]=\bx[n]-\bx_0$ be the error vector at time $n$. Subtracting $\bx_0$ from the left and right hand sides of (\ref{LMS}), using (\ref{lin_observation}) and relation $\mB\widetilde{\bx}[n]=\widetilde{\bx}[n]$, we obtain:
\begin{equation}
\label{error_recursion}
\widetilde{\bx}[n+1]=(\mI-\mu\,\mB\mD\mB)\,\widetilde{\bx}[n]+\mu\,\mB\mD \bv[n].
\end{equation}
Applying a GFT to each side of (\ref{error_recursion}) (i.e., multiplying by $\mU^H$), and exploiting the structure of matrix $\mB$ in (\ref{lowpass_operator}), we obtain
\begin{equation}
\label{error_recursion2}
\widetilde{\bs}[n+1]=(\mI-\mu\,\mathbf{\Sigma}\mU^H\mD\mU\mathbf{\Sigma})\,\widetilde{\bs}[n]+\mu\,\mathbf{\Sigma}\mU^H\mD \bv[n],
\end{equation}
where $\widetilde{\bs}[n]=\mU^H\widetilde{\bx}[n]$ is the GFT of the error $\widetilde{\bx}[n]$. From (\ref{error_recursion2}) and the definition of $\mathbf{\Sigma}$ in (\ref{lowpass_operator}), since $\widetilde{s}_i[n]=0$ for all $i \notin \F$, we can analyze the behavior of the error recursion (\ref{error_recursion2}) only on the support of $\widetilde{\bs}[n]$, i.e. $\widehat{\bs}[n]=\{\widetilde{s}_i[n], i\in \F\}\in \mathbb{C}^{|\F|}$. Thus, letting $\mU_{\mathcal{F}}\in \mathbb{C}^{N\times |\mathcal{F}|}$ be the matrix having as columns the eigenvectors of the Laplacian matrix associated to the frequency indices $\F$, the error recursion (\ref{error_recursion2}) can be rewritten in compact form as:
\begin{equation}
\label{error_recursion3}
\widehat{\bs}[n+1]=(\mI-\mu\,\mU_{\F}^H\mD\mU_{\F})\,\widehat{\bs}[n]+\mu\,\mU_{\F}^H\mD \bv[n].
\end{equation}
The evolution of the error $\widehat{\bs}[n]=\mU_{\mathcal{F}}^H\widetilde{\bx}[n]$ in the compact transformed domain is totally equivalent to the behavior of $\widetilde{\bx}[n]$ from a mean-square error point of view. Thus, using energy conservation arguments \cite{sayed2003energy}, we consider a general weighted squared error sequence $\|\widehat{\bs}[n]\|^2_{\boldsymbol{\Phi}}=\widehat{\bs}[n]^H\boldsymbol{\Phi}\widehat{\bs}[n]$, where $\boldsymbol{\Phi}\in \mathbb{C}^{|\F|\times|\F|}$ is any Hermitian nonnegative-definite matrix that we are free to choose. In the sequel, it will be clear the role played by a proper selection of the matrix $\boldsymbol{\Phi}$. Then, from (\ref{error_recursion3}) we can establish the following variance relation:
\begin{align}\label{var_relation}
\hspace{-.1cm}\mathbb{E}\|\widehat{\bs}[n+1]\|^2_{\boldsymbol{\Phi}}&=\mathbb{E}\|\widehat{\bs}[n]\|^2_{\boldsymbol{\Phi}'}
+\mu^2 \,\mathbb{E}\{\bv[n]^H\mD\mU_{\F}\boldsymbol{\Phi}\mU_{\F}^H\mD\bv[n]\} \nonumber\\
&\hspace{-.3cm}= \mathbb{E}\|\widehat{\bs}[n]\|^2_{\boldsymbol{\Phi}'}+\mu^2 \,{\rm Tr}(\boldsymbol{\Phi}\mU_{\F}^H\mD\mC_v\mD\mU_{\F})
\end{align}
where ${\rm Tr}(\cdot)$ denotes the trace operator, and
\begin{equation}
\label{Phi'}
\boldsymbol{\Phi}'=\left(\mI-\mu\,\mU_{\F}^H\mD\mU_{\F}\right)\boldsymbol{\Phi}\left(\mI-\mu\,\mU_{\F}^H\mD\mU_{\F}\right).
\end{equation}
Let $\boldsymbol{\varphi}={\rm vec}(\boldsymbol{\Phi})$ and $\boldsymbol{\varphi}'={\rm vec}(\boldsymbol{\Phi}')$, where the notation ${\rm vec}(\cdot)$ stacks the columns of $\boldsymbol{\Phi}$ on top of each other and ${\rm vec}^{-1}(\cdot)$ is the inverse operation. We
will use interchangeably the notation $\|\widehat{\bs}\|^2_{\boldsymbol{\Phi}}$ and $\|\widehat{\bs}\|^2_{\boldsymbol{\varphi}}$ to denote the same quantity $\widehat{\bs}^H\boldsymbol{\Phi}\widehat{\bs}$. Exploiting the Kronecker product property
$${\rm vec}(\mX\boldsymbol{\Phi}\mY)=(\mY^H\otimes\mX){\rm vec}(\boldsymbol{\Phi}),$$ and the trace property
$${\rm Tr}(\boldsymbol{\Phi}\mX)={\rm vec}(\mX^H)^T{\rm vec}(\boldsymbol{\Phi}),$$
in the relation (\ref{var_relation}), we obtain:
\begin{align}\label{var_relation2}
\mathbb{E}\|\widehat{\bs}[n+1]\|^2_{\boldsymbol{\varphi}}&=\mathbb{E}\|\widehat{\bs}[n]\|^2_{\boldsymbol{\mathbf{Q}\varphi}}
+\mu^2 {\rm vec}(\mG)^T\boldsymbol{\varphi}
\end{align}
where
\begin{align}
\mG&=\mU_{\F}^H\mD\mC_v \mD\mU_{\F}  \label{G}\\
\mQ&= (\mI-\mu\, \mU_{\F}^H\mD\mU_{\F})\otimes (\mI-\mu\, \mU_{\F}^H\mD\mU_{\F}). \label{Q}
\end{align}
The following theorem guarantees the asymptotic mean-square stability (i.e., convergence in the mean and mean-square error sense)
of the LMS algorithm in (\ref{LMS}). \smallskip
\textit{\begin{theorem}\label{theorem_stability}
Assume model (\ref{lin_observation}) holds. Then, for any bounded initial condition, the LMS strategy (\ref{LMS}) asymptotically converges in the mean-square error sense if the sampling operator $\mD$ and the step-size $\mu$ are chosen to satisfy (\ref{|DcB|<1}) and
\begin{equation}\label{step}
0< \mu < \frac{2}{\lambda_{\max}\left(\mU_{\F}^H\mD\mU_{\F}\right)},
\end{equation}
with $\lambda_{\max}(\mA)$ denoting the maximum eigenvalue of the symmetric matrix $\mA$. Furthermore, it follows that, for sufficiently small step-sizes:
\begin{equation}\label{lim_sup}
\lim_{n\rightarrow\infty}\sup_n\,\mathbb{E}\|\widehat{\bs}[n]\|^2 \,=\, O(\mu).
\end{equation}
\end{theorem}}

\begin{proof}
Letting $\br={\rm vec}(\mG)$, recursion (\ref{var_relation2}) can be equivalently recast as:
\begin{align}\label{var_relation3}
\mathbb{E}\|\widehat{\bs}[n]\|^2_{\boldsymbol{\varphi}}&=\mathbb{E}\|\widehat{\bs}[0]\|^2_{\boldsymbol{\mathbf{Q}}^n\boldsymbol{\varphi}}
+\mu^2 \br^T\sum_{l=0}^{n-1}\boldsymbol{\mathbf{Q}}^l\boldsymbol{\varphi}
\end{align}
where $\mathbb{E}\|\widehat{\bs}[0]\|$ denotes the initial condition. We first note that if $\mQ$ is stable, $\mQ^n\rightarrow \mathbf{0}$ as $n\rightarrow\infty$. In this way, the first term on the RHS of (\ref{var_relation3}) vanishes asymptotically. At the same time, the convergence of the second term on the RHS of (\ref{var_relation3}) depends only on the geometric series of matrices $\sum_{l=0}^{n-1}\boldsymbol{\mathbf{Q}}^l$, which is known to be convergent to a finite value if the matrix $\mathbf{Q}$ is a stable matrix \cite{Horn:1985:MA:5509}. In summary, if $\mQ$ is stable, the RHS of (\ref{var_relation3}) asymptotically converges to a finite value, and we conclude that $\mathbb{E}\|\widehat{\bs}[n]\|^2_{\boldsymbol{\varphi}}$ will converge to a steady-state value. From (\ref{Q}), we deduce that $\mQ$ is stable if matrix $\mI-\mu\mU_{\F}^H\mD\mU_{\F}$ is stable as well. This holds true under the two following conditions. The first condition is that matrix $\mU_{\F}^H\mD\mU_{\F}=\mI-\mU_{\F}^H\overline{\mD}\mU_{\F}$ must have full rank, i.e. (\ref{|DcB|<1}) holds true, where we have exploited the relation $$\|\overline{\mD}\mU_{\F}\|=\|\overline{\mD}\mU \boldsymbol{\Sigma}\|=\|\overline{\mD}\mB\|.$$
Now recalling that, for any Hermitian matrix $\mX$, it holds $\|\mX\|=\rho(\mX)$ \cite{Horn:1985:MA:5509}, with $\rho(\mX)$ denoting the spectral radius of $\mX$, the second condition guaranteing the stability of $\mQ$ is that
$\|\mI-\mu\mU_{\F}^H\mD\mU_{\F}\|<1$, which holds true for any step-sizes satisfying (\ref{step}). This concludes the first part of the Theorem.

We now prove the second part. Selecting $\boldsymbol{\varphi}={\rm vec}(\mI)$ in (\ref{var_relation3}), we obtain the following bound
\begin{align}\label{bound}
\mathbb{E}\|\widehat{\bs}[n]\|^2\leq  \mathbb{E}\|\widehat{\bs}[0]\|^2_{\boldsymbol{\mathbf{Q}}^n{\rm vec}(\mathbf{I})}
+\mu^2 c \sum_{l=0}^n\|\mQ\|^l
\end{align}
where $c=\|\br\|\|{\rm vec}(\mI)\|$. Taking the limit of (\ref{bound}) as $n\rightarrow\infty$, since $\|\mQ\|<1$ if conditions (\ref{|DcB|<1}) and (\ref{step}) hold, we obtain
\begin{align}\label{bound2}
\lim_{n\rightarrow\infty}\mathbb{E}\|\widehat{\bs}[n]\|^2\leq \frac{\mu^2 c}{1-\|\mQ\|}.
\end{align}
From (\ref{Q}), we have
\begin{align}\label{bound_norm_Q}
\|\mQ\|&=\|\mI-\mu\mU_{\F}^H\mD\mU_{\F}\|^2=\left(\rho\big(\mI-\mu\mU_{\F}^H\mD\mU_{\F}\big)\right)^2\nonumber\\
&\;\leq \;\max\left\{(1-\mu\delta)^2,(1-\mu\nu)^2\right\} \nonumber\\
&\;\stackrel{(a)}{\leq} \;1-2\mu\nu+\mu^2\delta^2
\end{align}
where $\delta=\lambda_{\max}(\mU_{\F}^H\mD\mU_{\F})$, $\nu=\lambda_{\min}(\mU_{\F}^H\mD\mU_{\F})$, and in (a) we have exploited $\delta\geq\nu$. Substituting (\ref{bound_norm_Q}) in (\ref{bound2}), we get
\begin{align}\label{bound3}
\lim_{n\rightarrow\infty}\mathbb{E}\|\widehat{\bs}[n]\|^2\leq \frac{\mu c}{2\nu-\mu\delta^2}.
\end{align}
It is easy to check that the upper bound (\ref{bound3}) does not exceed $\mu c/\nu$ for any stepsize $0<\mu<\nu/\delta^2$. Thus, we conclude that (\ref{lim_sup}) holds for sufficiently small step-sizes.
\end{proof}

\subsection{Steady-State Performance}

Taking the limit of (\ref{var_relation2}) as $n\rightarrow\infty$ (assuming conditions (\ref{|DcB|<1}) and (\ref{step}) hold true), we obtain:
\begin{align}\label{MSD2}
\lim_{n\rightarrow\infty}\mathbb{E}\|\widehat{\bs}[n]\|^2_{(\mathbf{I}-\mathbf{Q})\boldsymbol{\varphi}}=\mu^2 {\rm vec}(\mG)^T\boldsymbol{\varphi}.
\end{align}
Expression (\ref{MSD2}) is a useful result: it allows us to derive several performance metrics through the proper selection of
the free weighting parameter $\boldsymbol{\varphi}$ (or $\boldsymbol{\Phi}$). For instance, let us assume that one wants to evaluate the steady-state mean square deviation (MSD) of the LMS strategy in (\ref{LMS}). Thus, selecting $\boldsymbol{\varphi}=(\mathbf{I}-\mathbf{Q})^{-1}{\rm vec}(\mathbf{I})$ in (\ref{MSD2}), we obtain
\begin{align}\label{MSD}
\hspace{-.15cm}{\rm MSD}&=\lim_{n\rightarrow\infty} \mathbb{E}\|\widetilde{\bx}[n]\|^2 = \lim_{n\rightarrow\infty} \mathbb{E}\|\widehat{\bs}[n]\|^2 \nonumber\\
& =\mu^2{\rm vec}(\mG)^T (\mI-\mQ)^{-1}{\rm vec}(\mI).
\end{align}
If instead one is interested in evaluating the mean square deviation obtained by the LMS algorithm in (\ref{LMS}) when reconstructing the value of the signal associated to $k$-th vertex of the graph, selecting $\boldsymbol{\varphi}=(\mathbf{I}-\mathbf{Q})^{-1}{\rm vec}(\mathbf{U}_{\F}^H\mathbf{E}_k\mathbf{U}_{\F})$ in (\ref{MSD2}), we obtain
\begin{align}\label{MSD_k}
\hspace{-.15cm}{\rm MSD}_k=&\lim_{n\rightarrow\infty} \mathbb{E}\|\widetilde{\bx}[n]\|^2_{\mathbf{E}_k}= \lim_{n\rightarrow\infty} \mathbb{E}\|\widehat{\bs}[n]\|^2_{\mathbf{U}_{\F}^H\mathbf{E}_k\mathbf{U}_{\F}}\nonumber\\
=&\;\mu^2{\rm vec}(\mG)^T (\mI-\mQ)^{-1}{\rm vec}(\mathbf{U}_{\F}^H\mathbf{E}_k\mathbf{U}_{\F}),
\end{align}
where $\mathbf{E}_k={\rm diag}\{\bee_k\}$, with $\bee_k\in \mathbb{R}^N$ denoting the $k$-th canonical vector. In the sequel, we will confirm the validity of these theoretical expressions by comparing them with numerical simulations.

\subsection{Sampling Strategies}

As illustrated in the previous sections, the properties of the proposed LMS algorithm in (\ref{LMS}) strongly depend on the choice of the sampling set $\mathcal{S}$, i.e. on the vertex limiting operator $\mD$. Indeed, building on the previous analysis, it is clear that the sampling strategy must be carefully designed in order to: a) enable reconstruction of the signal; b) guarantee stability of the algorithm; and c) impose a desired mean-square error at convergence. In particular, we will see that, when sampling signals defined on graphs, besides choosing the right number of samples, whenever possible it is also fundamental to have a strategy indicating {\it where} to sample, as the samples' location plays a key role in the performance of the reconstruction algorithm in (\ref{LMS}). To select the best sampling strategy, one should optimize some performance criterion, e.g. the MSD in (\ref{MSD}), with respect to the sampling set $\S$, or, equivalently, the vertex limiting operator $\mD$. However, since this formulation translates inevitably into a selection problem, whose solution in general requires an exhaustive search over all the possible combinations, the complexity of such procedure becomes intractable also for graph signals of moderate dimensions. Thus, in the sequel we will provide some numerically efficient, albeit sub-optimal, greedy algorithms to tackle the problem of selecting the sampling set.

\begin{algorithm}[t]
\vspace{.1cm}
$\textit{Input Data}:$ $M$, the number of samples.

$\textit{Output Data}:$ $\S$, the sampling set.

$\textit{Function}:$ \hspace{.23cm} initialize $\S\equiv \emptyset$

\hspace{.1 cm} while $|\S|<M$

\hspace{0.3cm} $\displaystyle s=\arg \min_j \; {\rm vec}(\mG(\mD_{\S\cup\{j\}}))^T (\mI-\mQ(\mD_{\S\cup\{j\}}))^{\dag}{\rm vec}(\mI)$;


\hspace{0.3cm} $\S \leftarrow \S \cup \{s\}$;

\hspace{.1cm} end\vspace{.1cm}

\noindent \caption*{\label{alg:Greedy1}\textbf{Sampling strategy 1: Minimization of MSD}}
\end{algorithm}

\noindent \textbf{Greedy Selection - Minimum MSD}: This strategy aims at minimizing the MSD in (\ref{MSD}) via a greedy approach: the method iteratively selects the samples from the graph that lead to the largest reduction in terms of MSD. Since the proposed greedy approach starts from an initially empty sampling set, when $|\S|<|\F|$, matrix $\mI-\mQ$ in (\ref{MSD}) is inevitably rank deficient. Then, in this case, the criterion builds on the pseudo-inverse of the matrix $\mI-\mQ$ in (\ref{MSD}), denoted by $(\mI-\mQ)^{\dag}$, which coincides with the inverse as soon as $|\S|\geq|\F|$. The resulting algorithm is summarized in the table entitled ``Sampling strategy 1'', where we made explicit the dependence of matrices $\mG$ and $\mQ$ on the sampling operator $\mD$. In the sequel, we will refer to this method as the Min-MSD strategy.

\begin{algorithm}[t]

\vspace{.1cm}
$\textit{Input Data}:$ $M$, the number of samples.

$\textit{Output Data}:$ $\S$, the sampling set. \smallskip

$\textit{Function}:$ \hspace{.23cm} initialize $\S\equiv \emptyset$

\hspace{2 cm} while $|\S|<M$

\hspace{2.3cm} $\displaystyle s=\arg \max_j \;\; \left| \mU_{\mathcal{F}}^H\mD_{\S\cup\{j\}}\mU_{\mathcal{F}} \right|_+$;

\hspace{2.3cm} $\S \leftarrow \S \cup \{s\}$;

\hspace{2cm} end \vspace{.1cm}

\protect\caption*{\label{alg:Greedy2}\textbf{Sampling strategy 2: Maximization of $\left|\mU_{\mathcal{F}}^H\mD\mU_{\mathcal{F}}\right|_+$}}
\end{algorithm}

\noindent \textbf{Greedy Selection - Maximum $|\mU_{\mathcal{F}}^H\mD\mU_{\mathcal{F}}|_+$}: In this case, the strategy aims at maximizing the volume of the parallelepiped build with the selected rows of matrix $\mU_{\mathcal{F}}$. The algorithm starts including the row with the largest norm in $\mU_{\mathcal{F}}$, and then it adds, iteratively, the rows having the largest norm and, at the same time, are as orthogonal as possible to the vectors already in $\S$. The rationale underlying this strategy is to design a well suited basis for the graph signal that we want to estimate. This criterion coincides with the maximization of the the pseudo-determinant of the matrix $\mU_{\mathcal{F}}^H\mD\mU_{\mathcal{F}}$ (i.e. the product of all nonzero eigenvalues), which is denoted by $\left|\mU_{\mathcal{F}}^H\mD\mU_{\mathcal{F}}\right|_+$. In the sequel, we motivate the rationale underlying this strategy.
Let us consider the eigendecomposition $\mQ=\mV\boldsymbol{\Lambda}\mV^H$. From (\ref{MSD}), we obtain:
\begin{align}\label{MSD3}
{\rm MSD}&=\mu^2{\rm vec}(\mG)^T (\mI-\mQ)^{-1}{\rm vec}(\mI)\nonumber\\
&= \mu^2 {\rm vec}(\mG)^T \mV (\mI-\boldsymbol{\Lambda})^{-1} \mV^H {\rm vec}(\mI) \nonumber\\
&= \mu^2 \sum_{i=1}^{|\F|^2} \frac{p_i\cdot q_i}{1-\lambda_i(\mQ)}
\end{align}
where $\bp=\{p_i\}=\mV^H{\rm vec}(\mG)$, $\bq=\{q_i\}=\mV^H{\rm vec}(\mI)$.
From (\ref{MSD3}), we notice how the MSD of the LMS algorithm in (\ref{LMS}) strongly depends on the values assumed by the eigenvalues $\lambda_i(\mQ)$, $i=1,\ldots,|\F|^2$. In particular, we would like to design matrix $\mQ$ in (\ref{Q}) such that its eigenvalues are as far as possible from 1. From (\ref{Q}), it is easy to understand that
$$\lambda_i(\mQ)=\left(1-\mu\lambda_k(\mU_{\mathcal{F}}^H\mD\mU_{\mathcal{F}})\right)\left(1-\mu\lambda_l(\mU_{\mathcal{F}}^H\mD\mU_{\mathcal{F}})\right)$$
$k,l=1,\ldots,|\F|$. Thus, requiring $\lambda_i(\mQ)$, $i=1,\ldots,|\F|^2$, to be as far as possible from 1 translates in designing the matrix $\mU_{\mathcal{F}}^H\mD\mU_{\mathcal{F}}\in \mathbb{C}^{|\mathcal{F}|\times |\mathcal{F}|}$ such that its eigenvalues are as far as possible from zero. Thus, a possible surrogate criterion for the approximate minimization of (\ref{MSD3}) can be formulated as the selection of the sampling set $\mathcal{S}$ (i.e. operator $\mD$) that maximizes the determinant (i.e. the product of all eigenvalues) of the matrix $\mU_{\mathcal{F}}^H\mD\mU_{\mathcal{F}}$. When $|\S|<|\F|$, matrix $\mU_{\mathcal{F}}^H\mD\mU_{\mathcal{F}}$ is inevitably rank deficient, and the strategy builds on the pseudo-determinant of $\mU_{\mathcal{F}}^H\mD\mU_{\mathcal{F}}$. Of course, when $|\S|\geq|\F|$, the pseudo determinant coincides with the determinant. The resulting algorithm is summarized in the table entitled ``Sampling strategy 2''. In the sequel, we will refer to this method as the Max-Det sampling strategy.

\noindent \textbf{Greedy Selection - Maximum $\lambda^+_{\min}(\mU_{\mathcal{F}}^H\mD\mU_{\mathcal{F}})$}: Finally, using similar arguments as before, a further surrogate criterion for the minimization of (\ref{MSD3}) can be formulated as the maximization of the minimum nonzero eigenvalue of the matrix $\mU_{\mathcal{F}}^H\mD\mU_{\mathcal{F}}$, which is denoted by $\lambda_{\min}^+\left(\mU_{\mathcal{F}}^H\mD\mU_{\mathcal{F}}\right)$. This greedy strategy exploits the same idea of the sampling method introduced in \cite{chen2015discrete} in the case of batch signal reconstruction. The resulting algorithm is summarized in the table entitled ``Sampling strategy 3''. We will refer to this method as the Max-$\lambda_{\min}$ sampling strategy.

\begin{algorithm}[t]

\vspace{.1cm}
$\textit{Input Data}:$ $M$, the number of samples.

$\textit{Output Data}:$ $\S$, the sampling set. \smallskip

$\textit{Function}:$ \hspace{.23cm} initialize $\S\equiv \emptyset$

\hspace{2 cm} while $|\S|<M$

\hspace{2.3cm} $\displaystyle s=\arg \max_j \;\; \lambda_{\min}^+\left( \mU_{\mathcal{F}}^H\mD_{\S\cup\{j\}}\mU_{\mathcal{F}} \right)$;

\hspace{2.3cm} $\S \leftarrow \S \cup \{s\}$;

\hspace{2cm} end \vspace{.1cm}

\protect\caption*{\label{alg:Greedy3}\textbf{Sampling strategy 3: Maximization of $\lambda_{\min}^+\left( \mU_{\mathcal{F}}^H\mD\mU_{\mathcal{F}} \right)$}}
\end{algorithm}

In the sequel, we will illustrate some numerical results aimed at comparing the performance achieved by the proposed LMS algorithm using the aforementioned sampling strategies.

\subsection{Numerical Results}

In this section, we first illustrate some numerical results aimed at confirming the theoretical results in (\ref{MSD}) and (\ref{MSD_k}). Then, we will illustrate how the sampling strategy affects the performance of the proposed LMS algorithm in (\ref{LMS}). Finally, we will evaluate the effect of a graph mismatching in the performance of the proposed algorithm.

\begin{figure}[t]
\centering
\includegraphics[width=8cm]{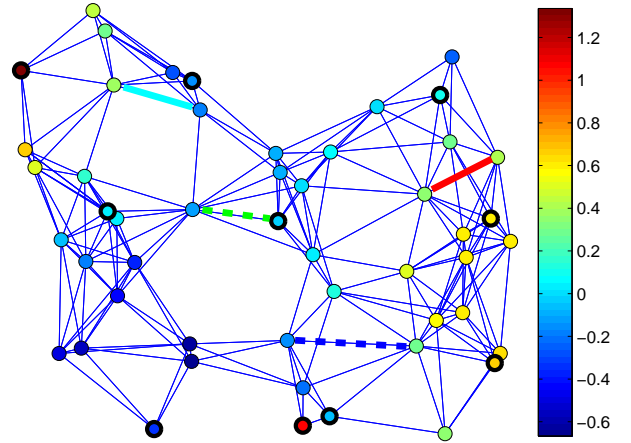}
\caption{Example of graph signal and sampling.}
\label{fig:Network}
\end{figure}

\begin{figure}[t]
\centering
\includegraphics[width=8cm]{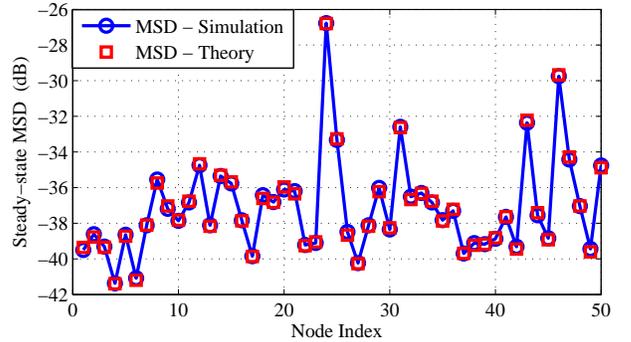}
\caption{Comparison between theoretical MSD in (\ref{MSD2}) and simulation results, at each vertex of the graph. The theoretical expressions match well with the numerical results.}
\label{fig:Theory}
\end{figure}

\noindent \textbf{Performance:} Let us consider the graph signal shown in Fig. \ref{fig:Network} and composed of $N=50$ nodes, where the color of each vertex denotes the value of the signal associated to it. The signal has a spectral content limited to the first ten eigenvectors of the Laplacian matrix of the graph in Fig. \ref{fig:Network}, i.e. $|\mathcal{F}|=10$. The observation noise in (\ref{lin_observation}) is zero-mean, Gaussian, with a diagonal covariance matrix, where each element is chosen uniformly random between 0 and 0.01. An example of graph sampling, obtained selecting $|\mathcal{S}|=10$ vertexes using the Max-Det sampling strategy, is also illustrated in Fig. \ref{fig:Network}, where the sampled vertexes have thicker marker edge. To validate the theoretical results in (\ref{MSD_k}), in Fig. \ref{fig:Theory} we report the behavior of the theoretical MSD values achieved at each vertex of the graph, comparing them with simulation results, obtained averaging over 200 independent simulations and 100 samples of squared error after convergence of the algorithm. The step-size is chosen equal to $\mu=0.5$ and, together with the selected sampling strategy $\mD$, they satisfy the reconstruction and stability conditions in (\ref{|DcB|<1}) and (\ref{step}). As we can notice from Fig. \ref{fig:Theory}, the theoretical predictions match well the simulation results.

\begin{figure}[t]
\centering
\includegraphics[width=8.1cm]{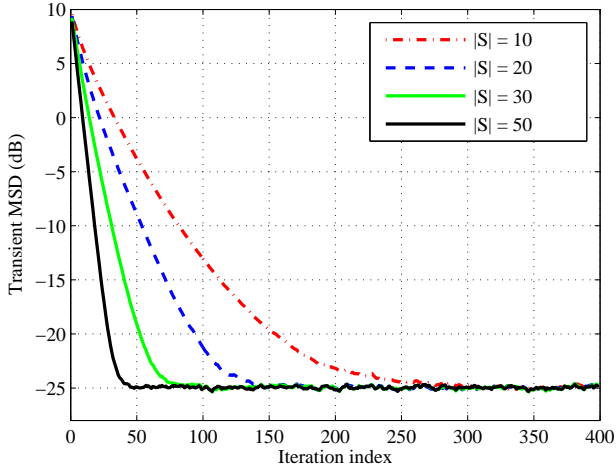}
\caption{Transient MSD, for different number of samples $|\mathcal{S}|$. Increasing the number of samples, the learning rate improves.}
\label{fig:Comp}
\end{figure}

\begin{figure}[t]
\centering
\includegraphics[width=8cm]{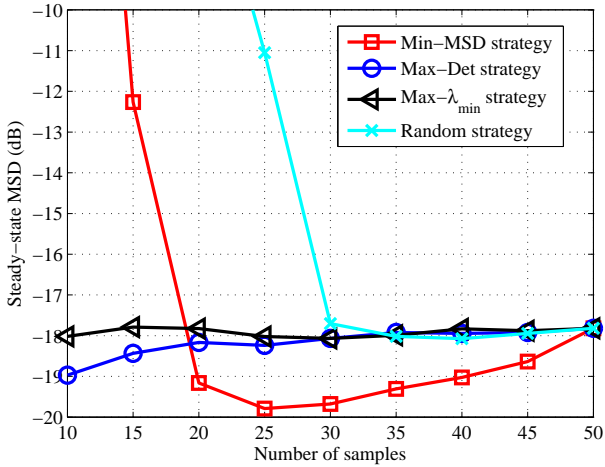}
\caption{Steady-state MSD versus number of samples, for different sampling strategies.}
\label{fig:Sampl_Comp}
\end{figure}
\noindent \textbf{Effect of sampling strategies:} It is fundamental to assess the performance of the LMS algorithm in (\ref{LMS}) with respect to the adopted sampling set $\mathcal{S}$. As a first example, using the Max-Det sampling strategy, in Fig. \ref{fig:Comp} we report the transient behavior of the MSD, considering different number of samples taken from the graph, i.e. different cardinalities $|\mathcal{S}|$ of the sampling set. The results are averaged over 200 independent simulations, and the step-sizes are tuned in order to have the same steady-state MSD for each value of $|\mathcal{S}|$. As expected, from Fig. \ref{fig:Comp} we notice how the learning rate of the algorithm improves by increasing the number of samples. Finally, in Fig. \ref{fig:Sampl_Comp} we illustrate the steady-state MSD of the LMS algorithm in (\ref{LMS}) comparing the performance obtained by four different sampling strategies, namely: a) the Max-Det strategy; b) the Max-$\lambda_{\min}$ strategy; c) the Min-MSD strategy; and d) the random sampling strategy, which simply picks at random $|\mathcal{S}|$ nodes. We consider the same parameter setting of the previous simulation.
The results are averaged over 200 independent simulations. As we can notice from Fig. \ref{fig:Sampl_Comp}, the LMS algorithm with random sampling can perform quite poorly, especially at low number of samples. This poor result of random sampling emphasizes that, when sampling a graph signal, what matters is not only the number of samples, but also (and most important) where the samples are taken. Comparing the other sampling strategies, we notice from Fig. \ref{fig:Sampl_Comp} that the Max-Det and Max-$\lambda_{\min}$ strategies perform well also at low number of samples ($|\mathcal{S}|=10$ is the minimum number of samples that allows signal reconstruction). As expected, the Max-Det strategy outperforms the Max-$\lambda_{\min}$ strategy, because it considers all the modes of the MSD in (\ref{MSD3}), as opposed to the single mode associated to the minimum eigenvalue considered by the Max-$\lambda_{\min}$ strategy. It is indeed remarkable that, for low number of samples, Max-Det outperforms also Min-MSD, even if the performance metric is MSD. There is no contradiction here because we need to remember that all the proposed methods are greedy strategies, so that there is no claim of optimality in all of them. However, as the number of samples increases above the limit $|\mathcal{S}|=|\mathcal{F}|=10$,  the Min-MSD strategy outperforms all other methods. This happens because the Min-MSD strategy takes into account information from {\it both} graph topology and spatial distribution of the observation noise (cf. (\ref{MSD})). Thus, when the number of samples is large enough to have sufficient degrees of freedom in selecting the samples' location, the Min-MSD strategy has the capability of selecting the vertexes in a good position to enable a well-conditioned signal recovery, with possibly low additive noise, thus improving the overall performance of the LMS algorithm in (\ref{LMS}). Conversely, when the number of samples is very close to its minimum value, the Min-MSD criterion may give rise to ill-conditioning of the signal recovery strategy because the low noise samples may be in sub-optimal positions with respect to signal recovery. This explains its losses with respect to Max-Det and Max-$\lambda_{min}$ strategies, for low values of the number of samples. This analysis suggests that an optimal design of the sampling strategy for graph signals should take into account processing complexity (in terms of number of samples), prior knowledge (e.g., graph structure, noise distribution), and achievable performance.

\begin{figure}[t]
\centering
\includegraphics[width=8cm]{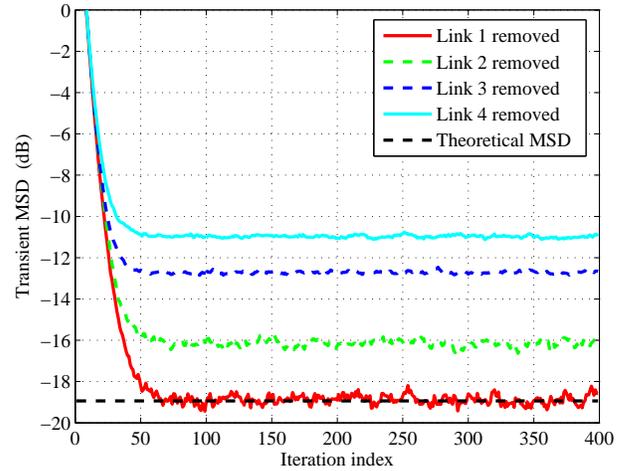}
\caption{Transient MSD versus iteration index, for different links removed from the original graph in Fig. \ref{fig:Network}.}
\label{fig:Graph_mismatch}
\end{figure}

\noindent \textbf{Effect of graph mismatching:} In this last example, we aim at illustrating how the performance of the proposed method is affected by a graph mismatching during the processing. To this aim, we take as a benchmark the graph signal in Fig. \ref{fig:Network}, where the signal bandwidth is set equal to $|\mathcal{F}|=10$. The bandwidth defines also the sampling operator $\mD$, which is selected through the Max-Det strategy, introduced in Sec. III.D, using $|\mathcal{S}|=10$ samples. Now, we assume that the LMS processing in (\ref{LMS}) is performed keeping fixed the sampling operator $\mD$, while adopting an operator $\mB$ in (\ref{lowpass_operator}) that uses the same bandwidth as for the benchmark case (i.e., the same matrix $\mathbf{\Sigma}_{\F}$), but different GFT operators $\mU$, which are generated as the eigenvectors of Laplacian matrices associated to graphs that differs from the benchmark in Fig. \ref{fig:Network} for one (removed) link. The aim of this simulation is to quantify the effect of a single link removal on the performance of the LMS strategy in (\ref{LMS}). Thus, in Fig. \ref{fig:Graph_mismatch}, we report the transient MSD versus the iteration index of the proposed LMS strategy, considering four different links that are removed from the original graph. The four removed links are those shown in Fig. \ref{fig:Network} using thicker lines; the colors and line styles are associated to the four behaviors of the transient MSD in Fig. \ref{fig:Graph_mismatch}.
The results are averaged over 100 independent simulations, using a step-size $\mu=0.5$. The theoretical performance in (\ref{MSD}) achieved by the ideal LMS, i.e. the one perfectly matched to the graph, are also reported as a benchmark. As we can see from Fig. \ref{fig:Graph_mismatch}, the removal of different links from the graph leads to very different performance obtained by the algorithm. Indeed, while removing Link 1 (i.e., the red one), the algorithm performs as in the ideal case, the removal of links 2, 3, and 4, progressively determine a worse performance loss. This happens because the structure of the eigenvectors of the Laplacian of the benchmark graph is more or less preserved by the removal of specific links. Some links have almost no effects (e.g., Link 1), whereas some others (e.g., Link 4) may lead to deep modification of the structure of such eigenvectors, thus determining the mismatching of the LMS strategy in (\ref{LMS}) and, consequently, its performance degradation. This example opens new theoretical questions that aim at understanding which links affect more the graph signals' estimation performance in situations where both the signal and the graph are jointly time-varying. We plan to tackle this exciting case in future work.



\section{LMS Estimation with Adaptive Graph Sampling}

The LMS strategy in (\ref{LMS}) assumes perfect knowledge of the support where the signal is defined in the graph frequency domain, i.e. $\mathcal{F}$. Indeed, this prior knowledge allows to define the projector operator $\mB$ in (\ref{lowpass_operator}) in a unique manner, and to implement the sampling strategies introduced in Sec. III.D. However, in many practical situations, this prior knowledge is unrealistic, due to the possible variability of the graph signal over time at various levels: the signal can be time varying according to a given model; the signal model may vary over time, for a given graph topology; the graph topology may vary as well over time. In all these situations, we cannot always assume that we have prior information about the frequency support $\mathcal{F}$, which must then be inferred directly from the streaming data $\by[n]$ in (\ref{lin_observation}). Here, we consider the important case where the graph is fixed, and the spectral content of the signal can vary over time in an unknown manner. Exploiting the definition of GFT in (\ref{GFT}), the signal observations in (\ref{lin_observation}) can be recast as:
\begin{align}
\label{lin_observation2}
\by[n]\,=\,&\mD\mU\bs_0+\mD\bv[n].
\end{align}
The problem then translates in estimating the coefficients of the GFT $\bs_0$, while identifying its support, i.e. the set of indexes where $\bs_0$ is different from zero. The support identification is deeply related to the selection of the sampling set. Thus, the overall problem can be formulated as the joint estimation of sparse representation $\bs$ and sampling strategy $\mD$ from the observations $\by[n]$ in (\ref{lin_observation2}), i.e.,
\begin{equation}
\label{Sparse_LMS_problem}
\min_{\bs,\mathbf{D}\in\mathcal{D}} \;\mathbb{E} \|\by[n]-\mD\mU\bs\|^2 +  \lambda\, f(\bs),
\end{equation}
where $\mathcal{D}$ is the (discrete) set that constraints the selection of the sampling strategy $\mD$, $f(\cdot)$ is a sparsifying penalty function (typically, $\ell_0$ or $\ell_1$ norms), and $\lambda>0$ is a parameter that regulates how sparse we want the optimal GFT vector $\bs$. Problem (\ref{Sparse_LMS_problem}) is a mixed integer nonconvex program, which is very complicated to solve, especially in the adaptive context considered in this paper. Thus, to favor low complexity online solutions for (\ref{Sparse_LMS_problem}), we propose an algorithm that alternates between the optimization of the vector $\bs$ and the selection of the sampling operator $\mD$. The rationale behind this choice is that, given an estimate for the support of vector $\bs$, i.e. $\F$, we can select the sampling operator $\mD$ in a very efficient manner through one of the sampling strategies illustrated in Sec. III.D. Then, starting from a random initialization for $\bs$ and a full sampling for $\mD$ (i.e., $\mD=\mI$), the algorithm iteratively proceeds as follows. First, fixing the value of the sampling operator $\mD[n]$ at time $n$, we update the estimate of the GFT vector $\bs$ using an online version of the celebrated ISTA algorithm \cite{daubechies2004iterative,beck2009fast}, which proceeds as:
\begin{equation}
\label{Threshold_LMS}
\bs[n+1]=\bT_{\lambda\mu}\left(\bs[n]+\mu\,\mU^H\mD[n]\left(\by[n]-\mU\bs[n]\right)\right),
\end{equation}
$n\geq0$, where $\mu>0$ is a (sufficiently small) step-size, and $\bT_{\gamma}(\bs)$ is a thresholding function that depends on the sparsity-inducing penalty $f(\cdot)$ in (\ref{Sparse_LMS_problem}). Several choices are possible, as we will illustrate in the sequel.
The aim of recursion (\ref{Threshold_LMS}) is to estimate the GFT $\bs_0$ of the graph signal $\bx_0$ in (\ref{lin_observation}), while selectively shrinking to zero all the components of $\bs_0$ that are outside its support, i.e., which do not belong to the bandwidth of the graph signal. Then, the online identification of the support of the GFT $\bs_0$ enables the adaptation of the sampling strategy, which can be updated using one of the strategies illustrated in Sec. III.D. Intuitively, the algorithm will increase (reduce) the number of samples used for the estimation, depending on the increment (reduction) of the current signal bandwidth. The main steps of the LMS algorithm with adaptive graph sampling are listed in Algorithm 2.

\begin{algorithm}[t]
\caption*{\textbf{Algorithm 2: LMS with Adaptive Graph Sampling}}
\vspace{.1cm}
Start with $\bs[0]$ chosen at random, $\mD[0]=\mI$, and $\mathcal{F}[0]=\mathcal{V}$. Given $\mu>0$, for each time $n>0$, repeat: \smallskip
\begin{enumerate}
  \item $\bs[n+1]=T_{\lambda\mu}\left(\bs[n]+\mu\,\mU^H\mD[n]\left(\by[n]-\mU\bs[n]\right)\right)$; \medskip
  \item Set $\mathcal{F}[n+1]=\{i\in\{1,\ldots,N\}:s_i[n+1]\neq0\}$; \medskip
  \item Given $\mU_{\mathcal{F}[n+1]}$, select $\mD[n+1]$ according to one of the criteria proposed in Sec. III.D;
\end{enumerate}\label{alg:AdaptiveLMS}\vspace{.1cm}
\end{algorithm}

\noindent \textbf{Thresholding functions :} Several different functions can be used to enforce sparsity. A commonly used thresholding
function comes directly by imposing an $\ell_1$ norm constraint in (\ref{Sparse_LMS_problem}), which is commonly known as the Lasso \cite{tibshirani1996regression}. In this case, the vector threshold function $\bT_\gamma(\bs)$ is the component-wise thresholding function $T_\gamma(s_m)$ applied to each element of vector $\bs$, with
\begin{align} \label{Lasso}
         T_\gamma(s_m)=\left\{
                      \begin{array}{ll}
                        s_m-\gamma, & \hbox{$s_m>\gamma$;} \\
                        0, & \hbox{$-\gamma\leq s_m \leq \gamma$;} \\
                        s_m+\gamma, & \hbox{$s_m<-\gamma$.}
                      \end{array}
                    \right.
\end{align}
The function $\bT_\gamma(\bs)$ in (\ref{Lasso}) tends to shrink all the components of the vector $\bs$ and, in particular, sets to zero the components whose magnitude are within the threshold $\gamma$. Since the Lasso constraint is known for introducing a large bias in the estimate, the performance would deteriorate for vectors that are not sufficiently sparse, i.e. graph signals with large bandwidth. To reduce the bias introduced by the Lasso constraint, several other thresholding functions can be adopted to improve the performance also in the case of less sparse systems. A potential improvement can be made by considering the non-negative Garotte estimator as in \cite{yuan2007non}, whose thresholding function is defined as a vector whose
entries are derived applying the threshold
\begin{align} \label{Garotte}
                    T_\gamma(s_m)=\left\{
                      \begin{array}{ll}
                        s_m\;(1-\gamma^2/s_m^2), \hspace{.5cm}& \hbox{$|s_m|>\gamma$;} \vspace{.2cm}\\
                        0, \hspace{.5cm}& \hbox{$|s_m|\leq\gamma$;}
                      \end{array}
                    \right.
\end{align}
$m=1,\ldots,M$. Finally, to completely remove the bias over the large components, we can implement a hard thresholding mechanism, whose function is defined as a vector whose entries are derived applying the threshold
\begin{align} \label{Hard_Threshold}
                    T_\gamma(s_m)=\left\{
                      \begin{array}{ll}
                        s_m, \hspace{.5cm}& \hbox{$|s_m|>\gamma$;} \vspace{.2cm}\\
                        0, \hspace{.5cm}&  \hbox{$|s_m|\leq\gamma$;}
                      \end{array}
                    \right.
\end{align}
In the sequel, numerical results will illustrate how different thresholding functions such as (\ref{Lasso}), (\ref{Garotte}), and (\ref{Hard_Threshold}), affect the performance of Algorithm 2.

\begin{figure}[t]
   \centering
   \includegraphics[width=8cm]{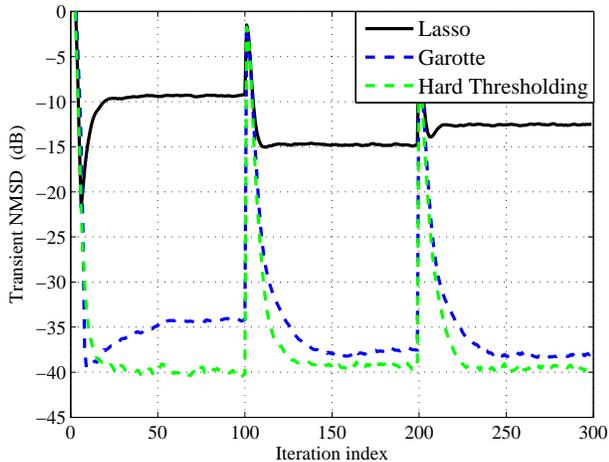}
   \caption{LMS with Adaptive Sampling: NMSD versus iteration index, for different thresholding functions.}\vspace{.2cm}
   \label{adaptive1}
\end{figure}

\begin{figure}[t]
   \centering
   \includegraphics[width=8cm]{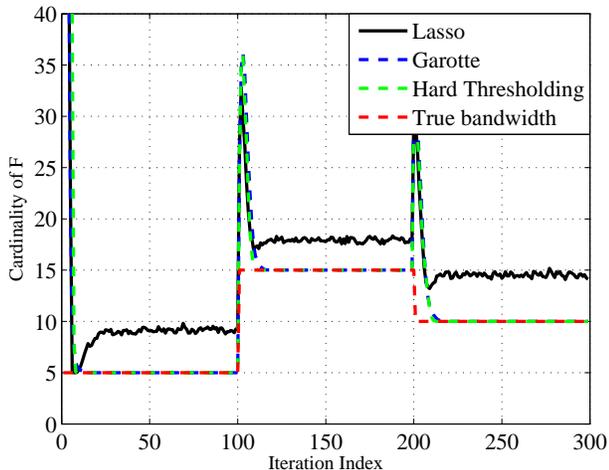}
   \caption{LMS with Adaptive Sampling: $|\mathcal{F}|$ versus iteration index, for different thresholding functions.}\vspace{.2cm}
   \label{adaptive2}
\end{figure}

\begin{figure}[t]
 \begin{minipage}[b]{8.5cm}
   \centering
   \includegraphics[width=7cm]{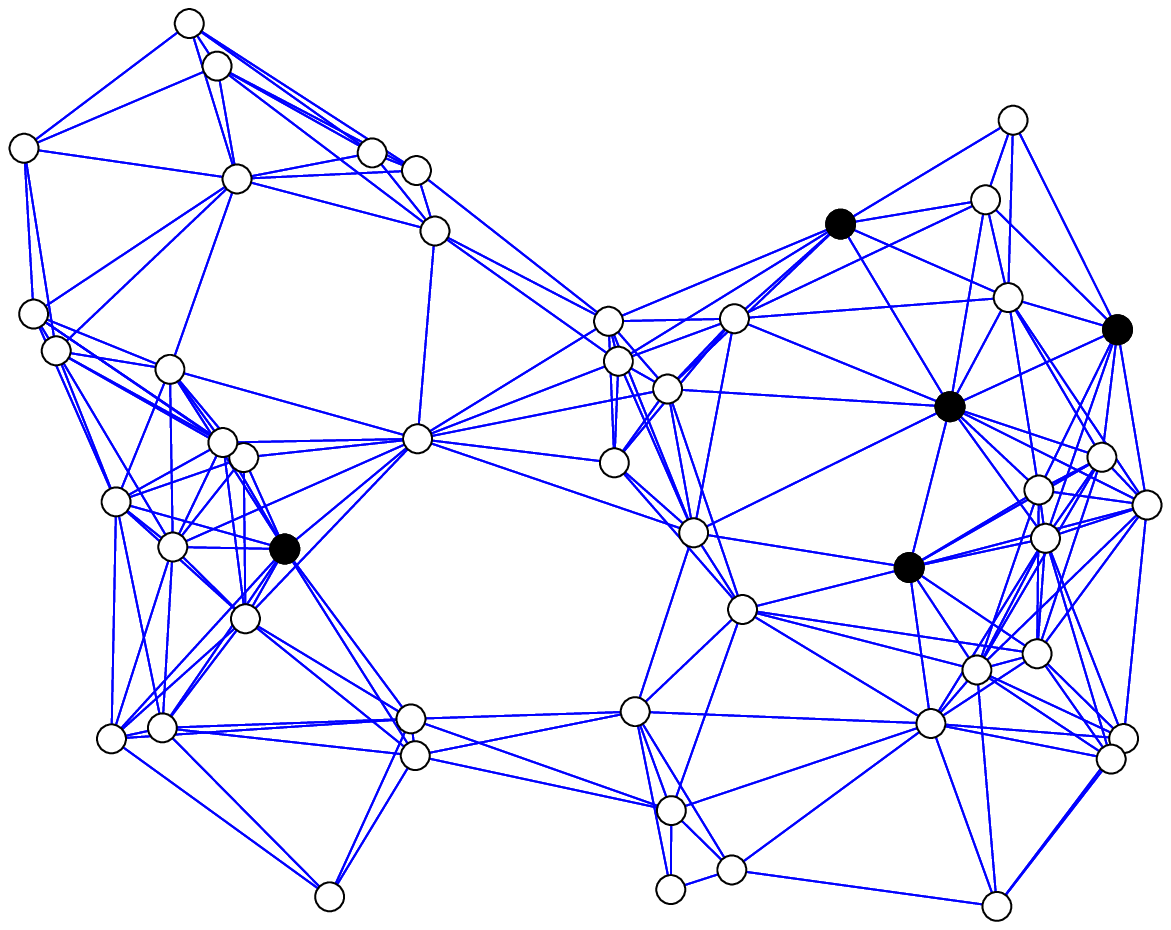}
   \caption{Optimal Sampling at iteration $n=80$.} \vspace{.4cm}
   \label{Net5}
 \end{minipage}
 \ \hspace{2mm} \hspace{3mm} \
 \begin{minipage}[b]{8.5cm}
  \centering
   \includegraphics[width=7cm]{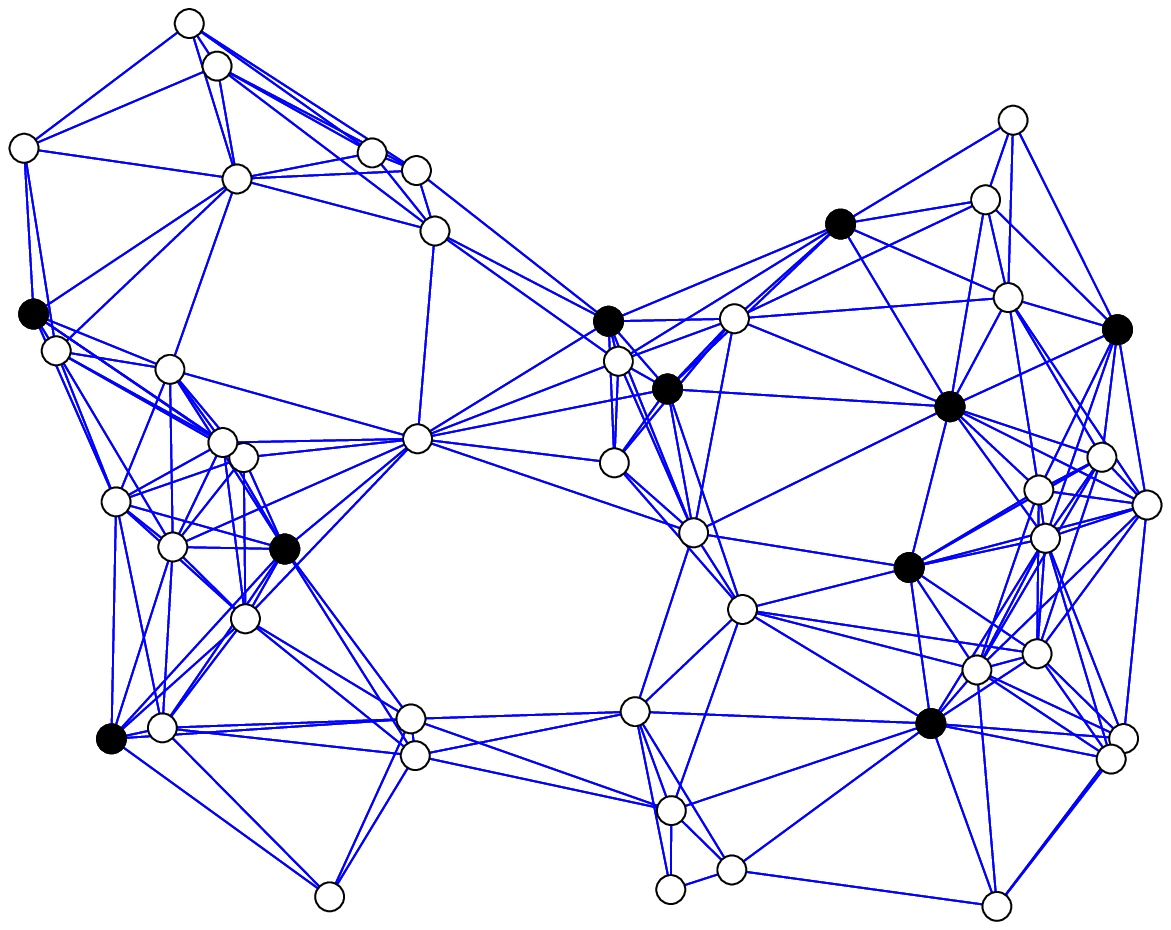}
   \caption{Optimal Sampling at iteration $n=180$.} \vspace{.4cm}
   \label{Net15}
 \end{minipage}
 \ \hspace{2mm} \hspace{3mm} \
 \begin{minipage}[b]{8.5cm}
  \centering
   \includegraphics[width=7cm]{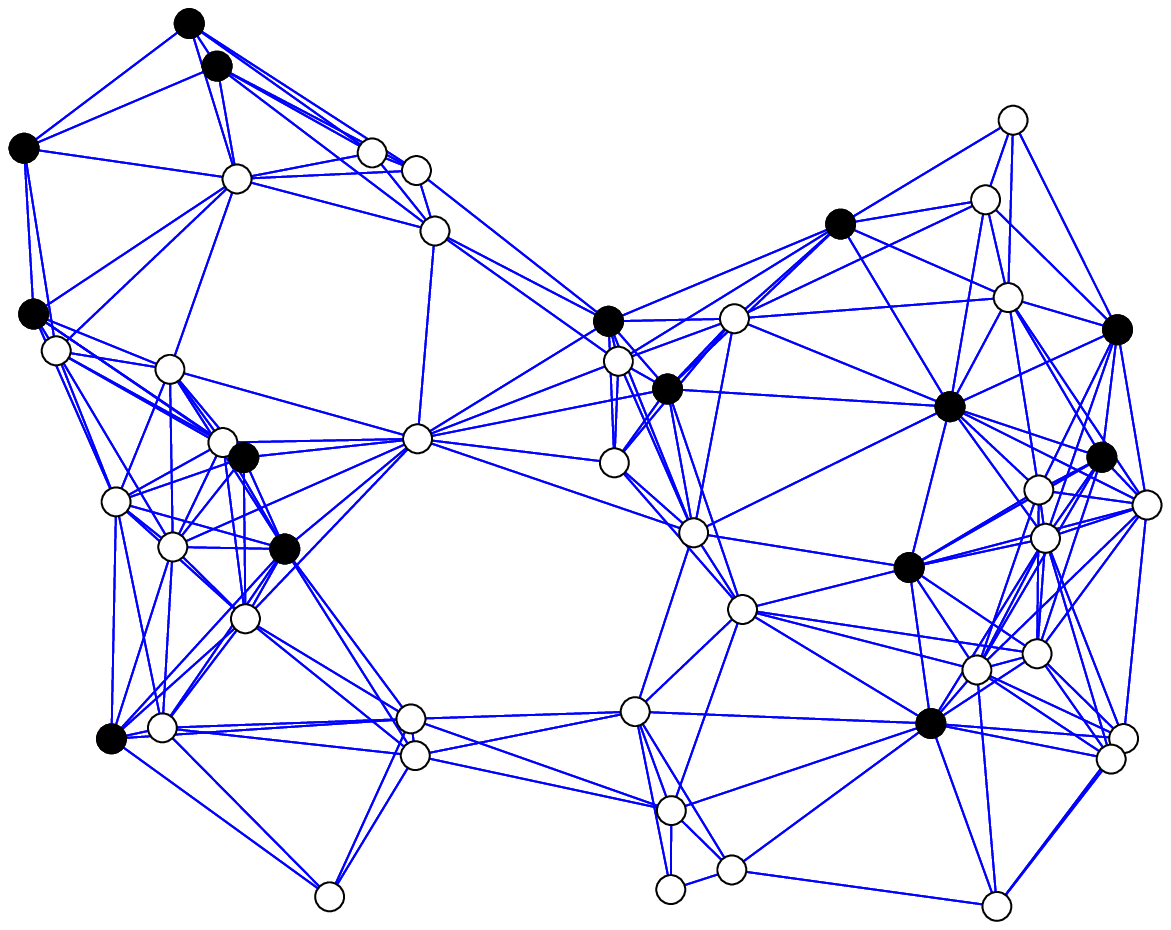}
   \caption{Optimal Sampling at iteration $n=280$.}
   \label{Net10}
 \end{minipage}
\end{figure}

\subsection*{Numerical Results}

In this section, we illustrate some numerical results aimed at assessing the performance of the proposed LMS method with adaptive graph sampling, i.e. Algorithm 2. In particular, to illustrate the adaptation capabilities of the algorithm, we simulate a scenario with a time-varying graph signal with $N=50$ nodes, which has the same topology shown in Fig. \ref{fig:Network}, and spectral content that switches between the first 5, 15, and 10 eigenvectors of the Laplacian matrix of the graph. The elements of the GFT $\bs_0$ inside the support are chosen to be equal to 1. The observation noise in (\ref{lin_observation}) is zero-mean, Gaussian, with a diagonal covariance matrix $\mC_v=\sigma_v^2 \mI$, with $\sigma_v^2=4\times 10^{-4}$. In Fig. \ref{adaptive1} we report the transient behavior of the normalized Mean-Square Deviation (NMSD), i.e.
$${\rm NMSD}[n]= \frac{\|\bs[n]-\bs_0\|^2}{\|\bs_0\|^2},$$
versus the iteration index, considering the evolution of Algorithm 2 with three different thresholding functions, namely: a) the Lasso threshold in (\ref{Lasso}), the Garotte threshold in (\ref{Garotte}), and the hard threshold in (\ref{Hard_Threshold}). Also, in Fig. \ref{adaptive2}, we illustrate the behavior of the estimate of the cardinality of $\mathcal{F}$ versus the iteration index (cf. Step 2 of  Algorithm 2), obtained by the three aforementioned strategies at each iteration. The value of the cardinality of $\mathcal{F}$ of the true underlying graph signal is also reported as a benchmark. The curves are averaged over 100 independent simulations. The step-size is chosen to be $\mu=0.5$, the sparsity parameter $\lambda=0.1$, and thus the threshold is equal to $\gamma=\mu\lambda=0.05$ for all strategies. The sampling strategy used in Step 3 of Algorithm 2 is the Max-Det method introduced in Sec. III.D, where the number of samples $M[n]$ to be selected at each iteration is chosen to be equal to the current estimate of cardinality of the set $F[n]$. As we can notice from Fig. \ref{adaptive1}, the LMS algorithm with adaptive graph sampling is able to track time-varying scenarios, and its performance is affected by the adopted thresholding function. In particular, from Fig. \ref{adaptive1}, we notice how the algorithm based on the hard thresholding function in (\ref{Hard_Threshold}) outperforms the other strategies in terms of steady-state NMSD, while having the same learning rate. The Garotte based algorithm has slightly worse performance with respect to the method exploiting hard thresholding, due to the residual bias introduced at large values by the thresholding function in (\ref{Garotte}). Finally, we can notice how the LMS algorithm based on Lasso may lead to very poor performance, due to misidentifications of the true graph bandwidth. This can be noticed from Fig. \ref{adaptive2} where,
while the Garotte and hard thresholding strategies are able to learn exactly the true bandwidth of the graph signal (thus leading to very good performance in terms of NMSD, see Fig. \ref{adaptive1}), the Lasso strategy overestimates the bandwidth of the signal, i.e. the cardinality of the set $\mathcal{F}$ (thus leading to poor estimation performance, see Fig. \ref{adaptive1}). Finally, to illustrate an example of adaptive sampling, in Figs. \ref{Net5}, \ref{Net15}, and \ref{Net10} we report the samples (depicted as black nodes) chosen by the proposed LMS algorithm based on hard thresholding at iterations $n=80$, $n=180$, and $n=280$. As we can notice from Figs. \ref{adaptive1}, \ref{adaptive2} and \ref{Net5}, \ref{Net15}, and \ref{Net10}, the algorithm always selects a number of samples equal to the current value of the signal bandwidth, while guaranteeing good reconstruction performance.


\section{Application to Power Spatial Density Estimation in Cognitive Networks}

The advent of intelligent networking of heterogeneous devices such as those deployed to monitor the 5G networks, power grid, transportation networks, and the Internet, will have a strong impact on the underlying systems. 
Situational awareness provided by such tools will be the key enabler for effective information dissemination, routing and congestion control, network health management, risk analysis, and security assurance. The vision is for ubiquitous smart network devices to enable data-driven statistical learning algorithms for distributed, robust, and online network operation and management, adaptable to the dynamically evolving network landscape with minimal need for human intervention. In this context, the unceasing demand for continuous situational awareness in cognitive radio (CR) networks calls for innovative signal processing algorithms, complemented by sensing platforms to accomplish the objectives of layered sensing and control. These challenges are embraced in the study of power cartography, where CRs collect data to estimate the distribution of power across space, namely the power spatial density (PSD). Knowing the PSD at any location allows CRs to dynamically implement a spatial reuse of idle bands. The estimated PSD map need not be extremely accurate, but precise enough to identify idle spatial regions.

In this section, we apply the proposed framework for LMS estimation of graph signals to spectrum cartography in cognitive networks. We consider a 5G scenario, where a dense deployment of radio access points (RAPs) is envisioned to provide a service environment characterized by very low latency and high rate access. Each RAP collects streaming data related to the spectrum utilization of primary users (PU's) at its geographical position. This information can then be sent to a processing center, which collects data from the entire system, through high speed wired links. The aim of the center is then to build a spatial map of the spectrum usage, while processing the received data on the fly and envisaging proper sampling techniques that enable a proactive sensing of the system from only a limited number of RAP's measurements. As we will see in the sequel, the proposed approach hinges on the graph structure of the signal received from the RAP's, thus enabling real-time PSD estimation from a small set of observations that are smartly sampled from the graph.

\begin{figure}[t]
\centering
\includegraphics[width=8cm]{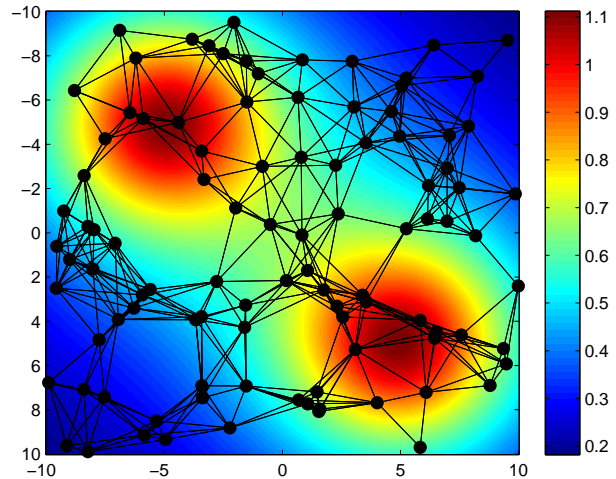}
\caption{PSD cartography: spatial distribution of primary users' power, small cell base stations deployment, graph topology, and graph signal.}
\label{fig:SS1}
\end{figure}
\noindent \textbf{Numerical examples:} Let us consider an operating region where 100 RAPs are randomly deployed to produce a map of the spatial distribution of power generated by the transmissions of two active primary users. The PU's emit electromagnetic radiation with power equal to 1 Watt. For simplicity, the propagation medium is supposed to introduce a free-space path loss attenuation on the PU's transmissions. The graph among RAPs is built from a distance based model, i.e. stations that are sufficiently close to each other are connected through a link (i.e. $a_{ij}=1$, if nodes $i$ and $j$ are neighbors). In Fig. \ref{fig:SS1}, we illustrate a pictorial description of the scenario, and of the resulting graph signal. We assume that each RAP is equipped with an energy detector, which estimates the received signal using 100 samples, considering an additive white Gaussian noise with variance $\sigma_v^2= 10^{-4}$. The resulting signal is not perfectly band-limited, but it turns out to be smooth over the graph, i.e. neighbor nodes observe similar values. This implies that sampling such signals inevitably introduces aliasing during the reconstruction process. However, even if we cannot find a limited (lower than $N$) set of frequencies where the signal is completely localized, the greatest part of the signal energy is concentrated at low frequencies. This means that if we process the data using a sufficient number of observations and (low) frequencies, we should still be able to reconstruct the signal with a satisfactory performance.

\begin{figure}[t]
\centering
\includegraphics[width=8cm]{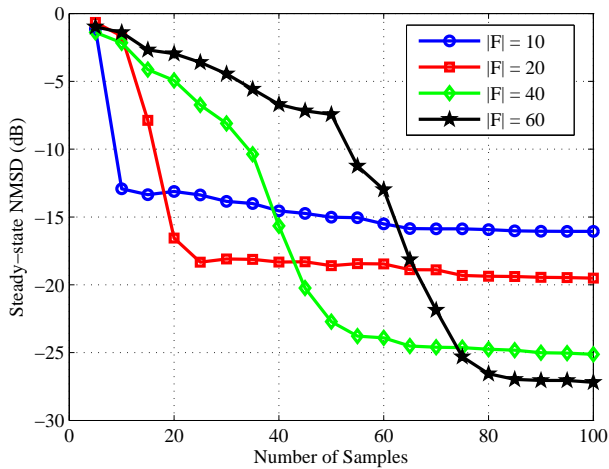}
\caption{PSD cartography: Steady-state NMSD versus number of samples taken from the graph, for different bandwidths used for processing.}
\label{fig:SS2}
\end{figure}

\begin{figure}[t]
\centering
\includegraphics[width=8cm]{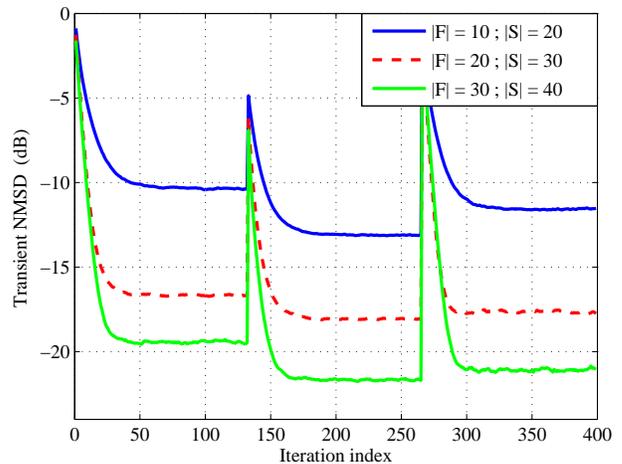}
\caption{PSD cartography: Transient NMSD versus iteration index, for different number of samples and bandwidths used for processing.}
\label{fig:SS3}
\end{figure}

To illustrate an example of cartography based on the LMS algorithm in (\ref{LMS}), in Fig. \ref{fig:SS2} we report the behavior of the steady-state NMSD versus the number of samples taken from the graph, for different bandwidths used for processing. The step-size is chosen equal to 0.5, while the adopted sampling strategy is the Max-Det method introduced in Sec. III.D. The results are averaged over 200 independent simulations. As expected, from Fig. \ref{fig:SS2}, we notice that the steady-state NMSD of the LMS algorithm in (\ref{LMS}) improves by increasing the number of samples and bandwidths used for processing. Interestingly, in Fig. \ref{fig:SS2} we can see a sort of threshold behavior: the NMSD is large for $|\S|<|\F|$, when the signal is undersampled, whereas the values become lower and stable as soon as $|\S|>|\F|$. Finally, we illustrate an example that shows the tracking capability of the proposed method in time-varying scenarios. In particular, we simulate a situation the two PU's switch between idle and active modes: for $0\leq n<133$ only the first PU transmits; for $133\leq n<266$ both PU's transmit; for $266\leq n\leq400$ only the second PU's transmits. In Fig. \ref{fig:SS3} we show the behavior of the transient NMSD versus iteration index, for different number of samples and bandwidths used for processing. The results are averaged over 200 independent simulations. From Fig. \ref{fig:SS3}, we can see how the proposed technique can track time-varying scenarios. Furthermore, its steady-state performance improves with increase in the number of samples and bandwidths used for processing. These results, together with those achieved in Fig. \ref{fig:SS2}, illustrate an existing tradeoff between complexity, i.e. number of samples used for processing, and mean-square performance of the proposed LMS strategy. In particular, using a larger bandwidth and a (consequent) larger number of samples for processing, the performance of the algorithm improves, at the price of a larger computational  complexity.

\section{Conclusions}

In this paper we have proposed LMS strategies for adaptive estimation of signals defined over graphs. The proposed strategies are able to exploit the underlying structure of the graph signal, which can be reconstructed from a limited number of observations properly sampled from a subset of vertexes, under a band-limited assumption. A detailed mean square analysis illustrates the deep connection between sampling strategy and the properties of the proposed LMS algorithm in terms of reconstruction capability, stability, and mean-square error performance. From this analysis, some sampling strategies for adaptive estimation of graph signals are also derived. Furthermore, to cope with time-varying scenarios, we also propose an LMS method with adaptive graph sampling, which estimates and tracks the signal support in the (graph)frequency domain, while at the same time adapting the graph sampling strategy. Several numerical simulations confirm the theoretical findings, and illustrate the potential advantages achieved by these strategies for adaptive estimation of band-limited graph signals. Finally, we apply the proposed method to estimate and track the spatial distribution of power transmitted by primary users in a cognitive network environment, thus illustrating the existing tradeoff between complexity and mean-square performance of the proposed strategy.

We expect that such processing tools will represent a key technology for the design and proactive sensing of Cyber Physical Systems, where a proper adaptive control mechanism requires the availability of data driven sampling strategies able to control the overall system by only checking a limited number of nodes, in order to collect correct information at the right time, in the right place, and for the right purpose.

\balance
\bibliographystyle{MyIEEE}
\bibliography{refs}

\end{document}